\documentclass[11pt]{article}

\usepackage[preprint]{acl}

\usepackage{times}
\usepackage{latexsym}

\usepackage[T1]{fontenc}

\usepackage[utf8]{inputenc}

\usepackage{microtype}

\usepackage{inconsolata}

\usepackage{graphicx}
\usepackage{amsmath} 
\usepackage{booktabs} 
\usepackage{enumitem}
\usepackage{times}  
\usepackage{helvet}  
\usepackage{courier}  
\usepackage{graphicx} 

\usepackage{algorithmic}
\usepackage{makecell}
\usepackage{booktabs}
\usepackage{multirow}
\usepackage{xcolor}
\usepackage{colortbl}
\usepackage{tcolorbox}
\usepackage[linesnumbered,ruled,vlined]{algorithm2e}
\usepackage{float}
\usepackage{amsmath, amssymb, amsfonts}

\title{DVPO: Distributional Value Modeling-based Policy Optimization for LLM Post-Training}

\author{
  \textbf{Dingwei Zhu$^{1}$}\thanks{\ \ Equal contribution.},
  \textbf{Zhiheng Xi$^{1*}$},
  \textbf{Shihan Dou$^1$},
  \textbf{Yuhui Wang$^1$},
  \textbf{Sixian Li$^1$}, \\
  \textbf{Junjie Ye$^1$},
  \textbf{Honglin Guo$^1$},
  \textbf{Shichun Liu$^1$},
  \textbf{Chenhao Huang$^1$},
  \textbf{Yajie Yang$^1$},\\
  \textbf{Junlin Shang$^1$},
  \textbf{Senjie Jin$^1$}, 
  \textbf{Ming Zhang$^1$},
  \textbf{Jiazheng Zhang$^1$},
  \textbf{Caishuang Huang$^1$},\\
  \textbf{Yunke Zhang$^2$},
  \textbf{Yuran Wang$^2$},
  \textbf{Tao Gui$^{1}$}\thanks{\ \ Corresponding author.}\\
  \vspace{0.1cm}
  \small $^1$College of Computer Science and Artificial Intelligence, Fudan University \\
  \small $^2$Honor Device Co., Ltd \\
  \texttt{\small dwzhu25@m.fudan.edu.cn, tgui@fudan.edu.cn}
}

\begin{document}
\maketitle

\begin{abstract}
Reinforcement learning (RL) has shown strong performance in LLM post-training, but real-world deployment often involves noisy or incomplete supervision. In such settings, complex and unreliable supervision signals can destabilize training and harm generalization. While existing approaches such as worst-case optimization (e.g., RFQI, CQL) and mean-based methods (e.g., PPO, GRPO) can improve stability, they often overlook generalization and may produce overly conservative policies, leading to uneven performance across diverse real scenarios. To this end, we introduce DVPO (Distributional Value Modeling with Risk-aware Policy Optimization), a new RL framework that combines conditional risk theory with distributional value modeling to better balance robustness and generalization. DVPO learns token-level value distributions to provide fine-grained supervision, and applies an asymmetric risk regularization to shape the distribution tails: it contracts the lower tail to dampen noisy negative deviations, while expanding the upper tail to preserve exploratory diversity. Across extensive experiments and analysis in multi-turn dialogue, math reasoning, and scientific QA, DVPO consistently outperforms PPO, GRPO, and robust Bellman-based PPO under noisy supervision, showing its potential for LLM post-training in the real-world.

\end{abstract}
\section{Introduction}
\begin{figure*}[h]
\centering
\includegraphics[width=1\linewidth]{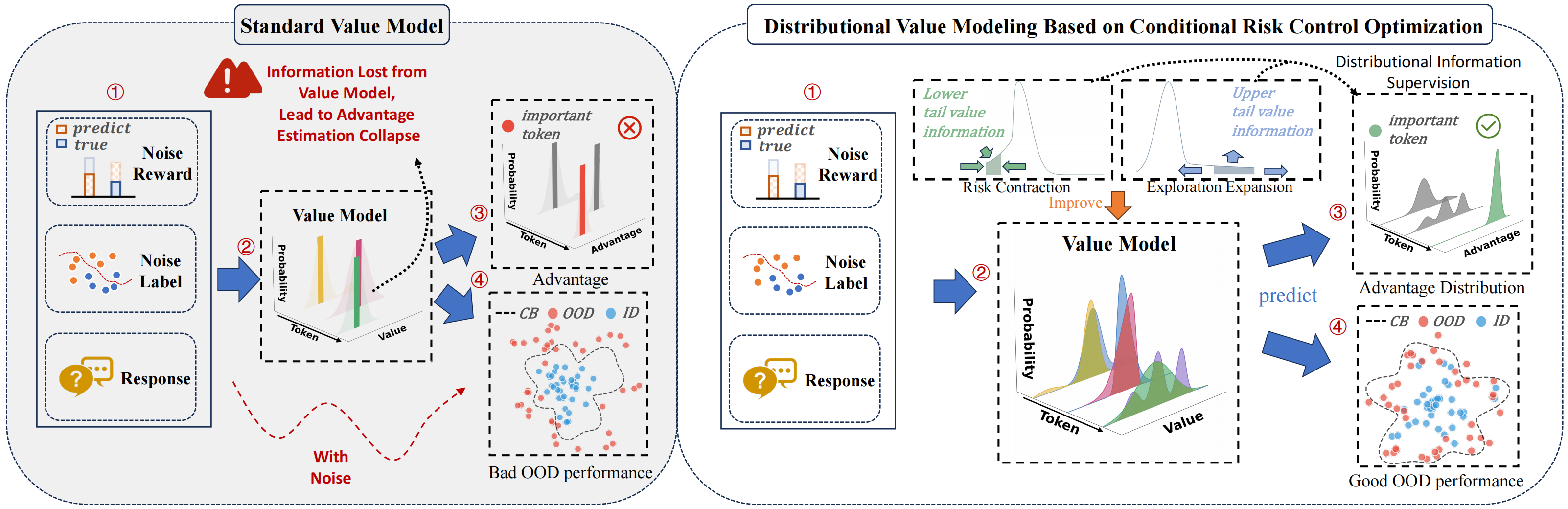}
\caption{Comparison between the Standard Value Model and our Distributional Value Model Based on Conditional Risk Control Policy Optimization (DVPO).
The standard value model  suffers from reward noise and biased value estimation, leading to unstable policy updates.
DVPO introduces a multi-head distributed value model to model the uncertainty of value, and further balances the model's learning of robustness and generalization in noise by means of lower-tail risk contraction and upper-tail exploratory expansion.}
\label{fig:all}
\end{figure*}

Reinforcement learning has achieved impressive results across diverse domains~\cite{xi2025agentgymrltrainingllmagents,xi2025bapostabilizingoffpolicyreinforcement,xi2024agentgymevolvinglargelanguage,xi2024traininglargelanguagemodels,ding2025mitigatingtailnarrowingllm,2024arXiv240219128C}. However, real-world RL often suffers from noisy or incomplete supervision, particularly when learning from human feedback or reward models. Such noise corrupts the value and advantage estimates, causing inaccurate policy updates and leading to unstable or divergent training dynamics. This issue is especially pronounced in LLM post-training~\cite{havrilla2024understandingeffectnoisellm,zhou2025qsharpprovablyoptimaldistributional} and autonomous agent learning~\cite{geng2024noisedistributiondecompositionbased,Para__2024}, where supervision signals are inherently complex and noisy.

Prior works have attempted to improve training stability through worst-case optimization, such as approaches based on robust Bellman operators~\cite{panaganti2022robustreinforcementlearningusing, kumar2020conservativeqlearningofflinereinforcement}. 
However, their pessimistic formulations often lead to overly conservative policies that suppress informative signals, thereby limiting the model’s ability to generalize advantage estimates to OOD scenarios. 
Other studies aim to control training variance via mean-based estimators, including GRPO and iterative value averaging~\cite{wang2023robustaveragerewardmarkovdecision}. 
While these methods help reduce instability, they still provide insufficient support for OOD generalization during training, often resulting in imbalanced advantage estimation performance across diverse real-world tasks. 
In practice, effective reinforcement learning methods must strike a balance between robustness to noisy rewards and cross-domain generalization ability~\cite{su2025crossingrewardbridgeexpanding}, enabling reliable performance under the diverse and challenging conditions encountered in real-world applications.

In this work, we propose a novel perspective to address this challenge by combining conditional risk theory~\cite{kisiala2015conditionalvalueatrisktheoryapplications} with a distributional training method~\cite{shi2023distributionallyrobustmodelbasedoffline}. Instead of purely minimizing worst-case outcomes or learning the expectation of future rewards, our goal is to maximize expected performance under controlled risk: we enhance generalization through more comprehensive distributional supervision and improve training robustness via distributional constraints.

We introduce \textbf{DVPO} (Distributional Value Modeling with Risk-aware Policy Optimization), a robust RL framework that predicts token-level value distributions instead of scalar estimates, enabling distributional advantage computation and fine-grained supervision. 
A conditional risk constraint asymmetrically regulates the distribution tails, contracting the lower tail to reduce noisy deviations, while expanding the upper tail to maintain exploratory diversity. 
This design preserves informative latent structures while optimizing expected returns under bounded risk, enabling appropriate advantage estimation across diverse scenarios and achieving a balance between robustness and generalization. 

Extensive experiments on multi-turn dialogue, mathematical reasoning, and scientific QA tasks show that DVPO consistently outperforms PPO, GRPO, and robust Bellman-based PPO variants under noisy supervision. 
By integrating distributional value learning with conditional risk control, DVPO achieves a robust and transferable learning framework, offering a scalable solution for real-world RL applications.
Our contributions are summarized as follows:
\begin{itemize}
    \item We propose DVPO, a robust RL framework with token-level value distributions for fine-grained supervision under noise.
    \item We design an asymmetric conditional risk constraint to regulate distribution tails, balancing robustness and generalization.
    \item We empirically demonstrate that DVPO outperforms representative baselines on multiple tasks, enhancing noise robustness and generalization.

\end{itemize}

\begin{figure*}[h]
\centering
\includegraphics[width=1\linewidth]{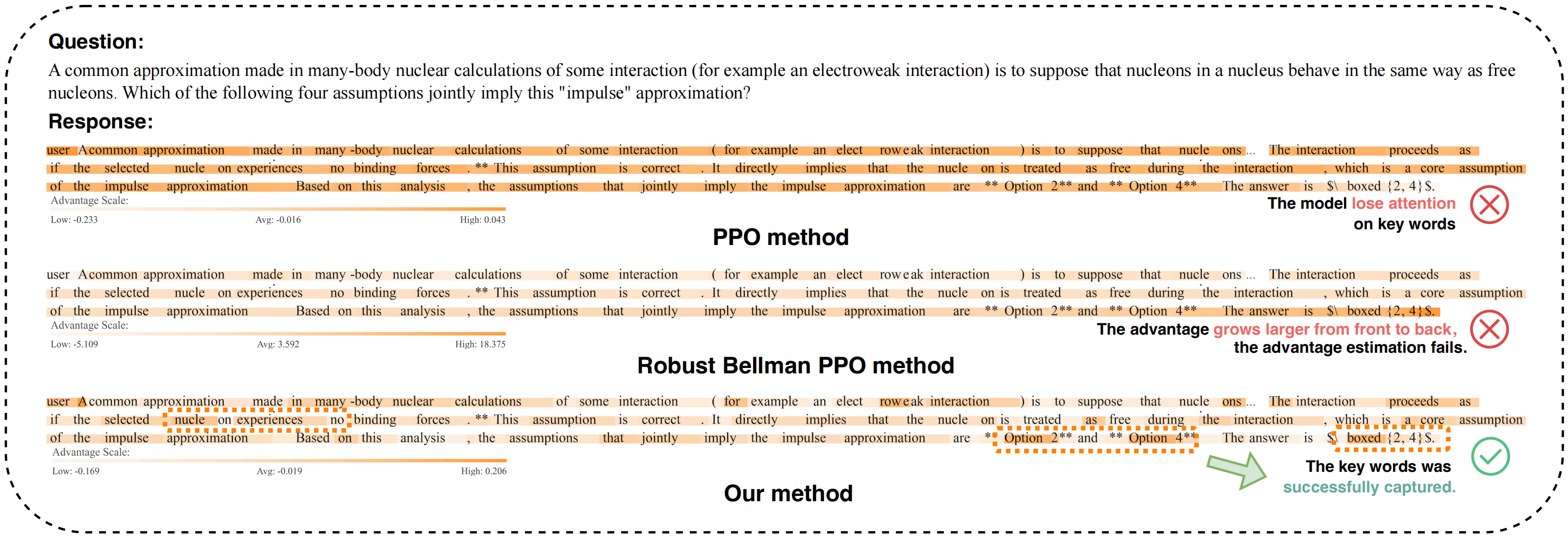}
\caption{Token-level advantage estimation for the same response across different methods. Our method exhibits sharper \textbf{focus on key words}.}
\label{fig:adv-vis}
\end{figure*}

\begin{figure*}[h]
\centering
\includegraphics[width=1\linewidth]{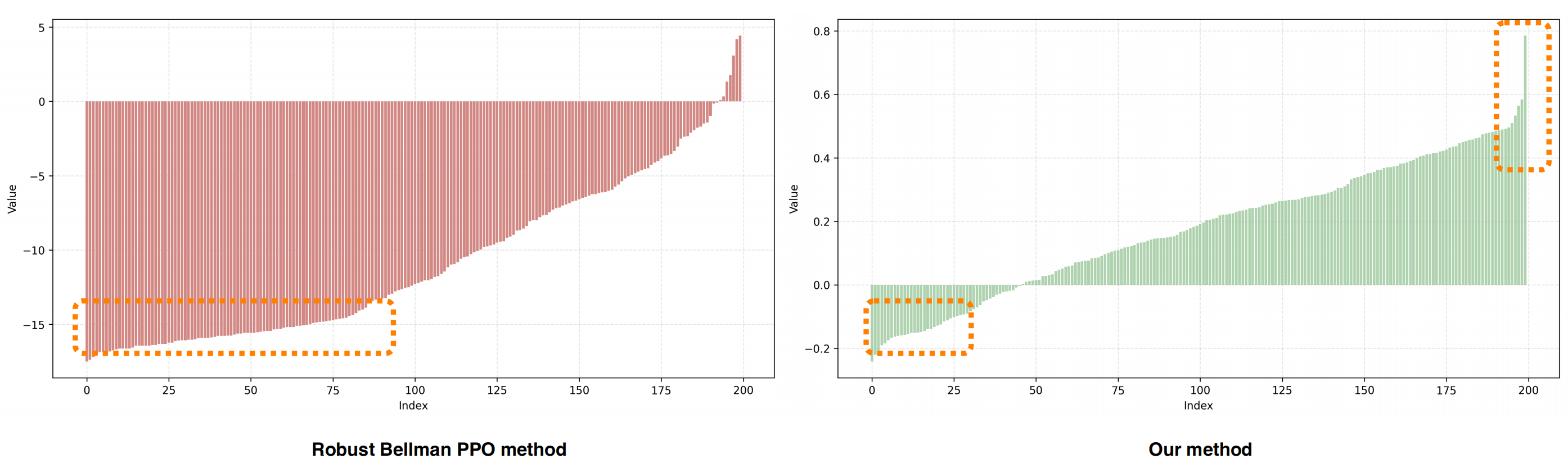}
\caption{Comparison of the output value distributions of the first-token method for the answer part. The robust Bellman PPO method contracts partially in the lower tail, with small variance during learning but insufficient exploration. Our method also contracts slightly in the lower tail, achieves significant exploratory expansion learning in the upper tail, and maintains a good balance between generalization and robustness.}
\label{fig:adv-fenbu}
\end{figure*}

\section{Related Work}

\paragraph{Robust Reinforcement Learning}
Robust Reinforcement Learning (RRL) aims to address the sensitivity of standard RL to model uncertainty, with Robust Markov Decision Processes (RMDPs)~\cite{Nilim2005RobustCO,bian2018continuoustimerobustdynamicprogramming} serving as a representative framework. RMDPs formulate decision-making as a zero-sum game and optimize worst-case performance through the robust Bellman operator~\cite{panaganti2022robustreinforcementlearningusing}, which introduces inner minimization into value updates to obtain pessimistic value estimates. Although such formulations guarantee convergence via contraction properties, their inherent pessimism often restricts exploration and leads to overly conservative policies, limiting generalization in complex environments.

To improve training stability, prior work has explored variance reduction via mean-based estimators such as GRPO~\cite{shao2024deepseekmathpushinglimitsmathematical,yu2025dapoopensourcellmreinforcement} and iterative value averaging~\cite{wang2023robustaveragerewardmarkovdecision}, as well as clipping-based strategies~\cite{becker2025trolltrustregionsimprove,yu2025dapoopensourcellmreinforcement}, including importance weight clipping~\cite{liu2025cpgdstablerulebasedreinforcement,minimax2025minimaxm1scalingtesttimecompute}. While these methods improve stability, they do not explicitly resolve the fundamental trade-off between robustness and OOD generalization.

In contrast, our work addresses this limitation by integrating distributional value modeling with conditional risk control, enabling more balanced learning that preserves robustness while improving generalization under noisy supervision.

\paragraph{Distributional Reinforcement Learning}Distributional Reinforcement Learning (DRL)~\cite{zanger2025diverseprojectionensemblesdistributional} extends traditional reinforcement learning by modeling the full probability distribution of future returns rather than merely their expectations. This distributional assumption provides richer learning signals, enabling the agent to capture higher-order statistics such as variance, skewness, and multimodality, thereby improving stability and robustness in decision-making. Early works include C51~\cite{bellemare2017distributionalperspectivereinforcementlearning}, which represents value distributions using discrete atoms; Quantile Regression DQN (QR-DQN)~\cite{dabney2017distributionalreinforcementlearningquantile}, which models the return distribution through quantile regression; and the Implicit Quantile Network (IQN)~\cite{dabney2018implicitquantilenetworksdistributional}, which continuously approximates the quantile function to enhance representational flexibility. Recently, in Reinforcement Learning from Human Feedback (RLHF), the Quantile Reward Model (QRM)~\cite{dorka2024quantileregressiondistributionalreward} extends reward modeling from scalar signals to full distributions, capturing the ambiguity and multidimensionality of human preferences and guiding optimization toward safer and more reliable behaviors. Q\#~\cite{zhou2025qsharpprovablyoptimaldistributional} further applies DRL to LLM post-training by learning the cumulative return distribution of the reference policy to compute a provably optimal KL-regularized guiding Q-function.
In this work, We build on these insights by introducing distributional supervision and conditional risk constraints into value modeling, enhancing stability and generalization under noisy supervision.

\section{Method}
\subsection{Motivation}

Training RL models in real-world environments inevitably involves noisy, incomplete supervision, especially in LLM post-training and autonomous agent learning. Existing robust RL methods, including robust Bellman operators \cite{panaganti2022robustreinforcementlearningusing} and mean estimation techniques (GRPO \cite{shao2024deepseekmathpushinglimitsmathematical}, PPO), iterative value averaging \cite{wang2023robustaveragerewardmarkovdecision}), fail to balance robustness and generalization. Robust Bellman operators enhance stability via worst-case optimization but yield overly conservative policies; mean estimation reduces variance through scalar-based biased estimation yet lacks cross-domain adaptability mechanisms. Both are inadequate for real-world RL scenarios that demand simultaneous resilience to noise and effective generalization.

A core limitation lies in their information-constrained value learning. Robust Bellman operators, in particular, suffer from excessively pessimistic estimation. As shown in Figure~\ref{fig:adv-fenbu}, training with minimum-value targets collapses the value distribution to its lower bound, producing stable but overly concentrated predictions. While this reduces variance, it suppresses high-value signals critical for diverse, informative feature learning. Under noisy supervision (Figure~\ref{fig:adv-vis}), this imbalance makes models overlook key words, letting spurious noise dominate task-related advantages.

To address these issues, we reformulate value learning from scalar prediction to distributional estimation, which captures richer value uncertainty and tail behaviors beyond biased mean or pessimistic estimates. 
Building on conditional risk control theory~\cite{kisiala2015conditionalvalueatrisktheoryapplications}, we impose asymmetric constraints on the value distribution by contracting the lower tail to suppress noise and risk, while relaxing the upper tail to preserve informative high-value signals and encourage exploration. 
This design enables more accurate advantage estimation across diverse scenarios and effectively balances robustness and generalization under noisy supervision.


\subsection{Distributional Architecture for Robust Value Representation}

\paragraph{Multi-Headed Quantile Ensemble}
To obtain a stable and noise-resilient value distribution under imperfect supervision, our critic adopts a multi-headed quantile ensemble. Instead of relying on a single parametric estimator, which is often sensitive to idiosyncratic noise and induces high variance in distributional RL, we use an ensemble of $N$ quantile heads to produce a more reliable distributional approximation.
A shared backbone network $h_\psi(s)$ first encodes the state into a latent representation. This representation is then passed to $N$ lightweight and independent quantile head networks, denoted by $\{f_{\phi_{Q,i}}\}_{i=1}^N$. Each head $i$ predicts a set of $M$ quantile values $\{\theta_{i,j}(s,a)\}_{j=1}^{M},$corresponding to a predefined grid of quantile levels $\mathcal{T} = \{\hat{\tau}_j = j /M\}_{j=1}^{M}.$
The final ensemble quantile function, $\hat{F}^{-1}_{Z(s,a)}$, is formed by averaging the outputs, creating a robust distributional baseline that is less susceptible to outlier estimates:
\begin{equation} \small \label{eq:ensemble_quantile}
\theta_j^{\text{ens}}(s,a) \triangleq \hat{F}^{-1}_{Z(s,a)}(\hat{\tau}_j) = \frac{1}{N} \sum_{i=1}^{N} \theta_{i,j}(s,a).
\end{equation}


\paragraph{Distributional Generalized Advantage Estimation}
We propagate this robust distributional information through time using a distributional extension of the Generalized Advantage Estimation~\cite{schulman2018highdimensionalcontinuouscontrolusing} framework. Our procedure operates entirely in the quantile domain to preserve the full spectrum of uncertainty during credit assignment. Let the value distribution at time $t$ be represented by its quantile vector $\Theta_{V_t}$. The distributional advantage, $A_t^D$, represented by its quantile vector $\Theta_{A_t}$, is then computed via a recursive update that lifts the standard GAE formulation to the quantile space:
\begin{equation}\small \label{eq:dgae_recurrence}
    \Theta_{A_t} = (r_t + \gamma \Theta_{V_{t+1}} - \Theta_{V_t}) + (\gamma \lambda) \Theta_{A_{t+1}}.
\end{equation}
The term in parentheses represents the distributional one-step TD-error. This recursive computation yields two critical outputs: a target return distribution for the value model update, defined by its quantile vector $\Theta_{Z'_t} = \Theta_{V_t} + \Theta_{A_t}$; and a scalar advantage estimate for the actor update, recovered by taking the expectation, $A(s_t, a_t) = \mathbf{E}[A_t^D] \approx \frac{1}{M} \sum_{j=1}^M \Theta_{A_t, j}$. By performing credit assignment entirely on distributions, this method provides a richer and more nuanced learning signal for our subsequent risk-aware value model optimization.

\subsection{Calibrating Risk-Aware Distributional Optimization}

Learning reliable value distributions under noisy or imperfect supervision requires simultaneously promoting robustness to adverse perturbations and preserving generalization to high-quality signals. To achieve this balance, we formulate a composite critic objective that shapes the predicted return distribution along multiple dimensions, including central quantile alignment, tail calibration, curvature control, and cross-distribution coherence.

Let the critic output a parametric quantile distribution 
$\hat{Z}(s,a)$ represented by $M$ quantile values 
$\Theta = \{\theta_j(s,a)\}_{j=1}^{M}$, evaluated at quantile levels 
$\mathcal{T} = \{\hat{\tau}_j\}_{j=1}^{M}$ where 
$\hat{\tau}_j = j /M$.
Let the target return distribution be 
$Z'(s,a)$ with corresponding quantiles 
$\Theta' = \{\theta'_j(s,a)\}_{j=1}^{M}$.
We denote by $\pi$ the permutation that sorts target quantiles, such that 
$\theta'_{\pi(1)} \le \dots \le \theta'_{\pi(M)}$.
For convenience, we define prediction errors 
$u_j = \theta'_j - \theta_j$.
E represents expectation. 
\paragraph{1. Core Quantile Regression}
The foundational component is the quantile Huber regression loss, which aligns each predicted quantile with its target counterpart:
\begin{equation}\small
\label{eq:qr_loss_new}
\mathcal{L}_{\text{QR}}
=
\mathbf{E}_{s,a}
\left[
\frac{1}{M}
\sum_{j=1}^{M}
\left|
\hat{\tau}_j - \mathbf{I}(u_j < 0)
\right|
L_\delta(u_j)
\right]
\end{equation}
where $L_\delta(\cdot)$ denotes the Huber loss and $\mathbf{I}(\cdot)$ is the indicator function.
This objective provides stable fitting but does not control robustness or generalization on its own.

\paragraph{2. Risk-Sensitive Quantile Weighting}
To explicitly enhance robustness to noisy negative rewards, we modulate quantile errors using a weighting function $w_{\text{risk}}(\hat{\tau}_j)$, parameterized by a risk-aversion coefficient $\gamma$:
\begin{equation}\small
\label{eq:risk_loss_new}
\mathcal{L}_{\text{Risk}}
=
\mathbf{E}_{s,a}
\left[
\frac{1}{M}
\sum_{j=1}^{M}
w_{\text{risk}}(\hat{\tau}_j)
\left|
\hat{\tau}_j - \mathbf{I}(u_j < 0)
\right|
L_\delta(u_j)
\right]
\end{equation}
Larger $\gamma$ increases emphasis on lower quantiles, promoting conservative updates that mitigate overfitting to noisy supervision.

\paragraph{3. Tail Expectation Calibration}
To regulate extreme outcomes, we align the expectations of both lower and upper tails.  
Let $K_{\alpha}=\lfloor \alpha M \rfloor$ denote the lower tail size and
$K_{\beta}=\lfloor \beta M \rfloor$ the upper tail size.

The CVaR-style objective stabilizes the worst-case region:
\begin{equation}\small
\label{eq:cvar_loss_new}
\mathcal{L}_{\text{CVaR}}
=
\mathbf{E}_{s,a}
\left[
\frac{1}{K_{\alpha}}
\left(
\sum_{j=1}^{K_{\alpha}}
\theta'_{\pi(j)}
-
\sum_{j=1}^{K_{\alpha}}
\theta_{\pi(j)}
\right)^2
\right]
\end{equation}

The upper-gain objective encourages accurate modeling of high-value outcomes, thus supporting generalization:
\begin{equation}\small
\mathcal{L}_{\text{Gain}} = \mathbf{E}_{s,a}
\left[\frac{1}{K_{\beta}}
\left(
\sum_{\substack{j=M \\ -K_{\beta}+1}}^{M}
\theta'_{\pi(j)}
-
\sum_{\substack{j=M \\ -K_{\beta}+1}}^{M}
\theta_{\pi(j)}
\right)^2
\right]
\label{eq:gain_loss_new}
\end{equation}

\paragraph{4. Controlling Systematic Bias via Mean-Shift Penalization}
To prevent persistent underestimation of the return distribution, we penalize cases where the predicted mean falls below the target mean:
\begin{equation}\small
\label{eq:shift_loss}
\mathcal{L}_{\text{Shift}}
=
\mathbf{E}_{s,a}
\left[
\text{ReLU}\!\left(
\mathbf{E}_j[\theta'_j] - \mathbf{E}_j[\theta_j]
\right)
\right]
\end{equation}
This reduces overly conservative value estimates that would otherwise impair exploration and generalization.

\paragraph{5. Tail-Shape Regularization}
We regulate dispersion in the extreme tails using variance-based penalties.  
Let $\mathrm{Var}(\Theta;\mathcal{I})$ denote the variance over index set $\mathcal{I}$, and define  
$\mathcal{I}_{\alpha} = \{1,\dots,K_{\alpha}\}$  
and  
$\mathcal{I}_{\beta}  = \{M-K_{\beta}+1,\dots,M\}$.

The tail-shape loss is:
\begin{equation}\small
\begin{split}
\mathcal{L}_{\text{Shape}} = \mathbf{E}_{s,a} \Big[ \text{ReLU}\!\big( \mathrm{Var}(\Theta;\mathcal{I}_{\alpha}) - \mathrm{Var}(\Theta';\mathcal{I}_{\alpha}) \big) \\
+ \text{ReLU}\!\big( \mathrm{Var}(\Theta';\mathcal{I}_{\beta}) - \mathrm{Var}(\Theta;\mathcal{I}_{\beta}) \big) \Big]
\end{split}
\label{eq:shape_loss}
\end{equation}
The first promotes a compact lower tail robustness,  
while the second encourages a sufficiently expressive upper tail generalization.

\paragraph{6. Tail Curvature Regularization}
To further shape extreme quantile geometry, we regularize discrete curvature using the second-order finite difference  
$\Delta^2 \theta_j = \theta_{j+1} - 2\theta_j + \theta_{j-1}$:
\begin{equation}\small
\begin{split}
\mathcal{L}_{\text{Curv}} = \mathbf{E}_{s,a} \left[ \text{ReLU}\!\big( \mathbf{E}_{j \in \mathcal{I}_{\alpha}}[\Delta^2 \theta_j] \big) \right. \\
\left. + \text{ReLU}\!\big( -\mathbf{E}_{j \in \mathcal{I}_{\beta}}[\Delta^2 \theta_j] \big) \right]
\end{split}
\label{eq:curvature_loss}
\end{equation}
This favors concave lower-tail curvature risk aversion  
and convex upper-tail curvature optimistic generalization.

\paragraph{7. Multi-Distribution Consistency Across Ensemble Heads}
If the critic uses multiple independently parameterized quantile distributions 
$\{\hat{Z}^{(i)}(s,a)\}_{i=1}^{N_{\text{dist}}}$,  
each producing quantiles 
$\Theta^{(i)} = \{\theta^{(i)}_j\}_{j=1}^{M}$,  
we encourage agreement across heads:
\begin{equation}\small
\label{eq:consistency_loss}
\mathcal{L}_{\text{Consist}}
=
\frac{1}{N_{\text{pair}}}
\sum_{i<j}
\mathbf{E}_{s,a}
\big[
\| \Theta^{(i)} - \Theta^{(j)} \|_2^2
\big]
\end{equation}
where $N_{\text{pair}}$ counts all head pairs.

\paragraph{Final Composite Critic Loss}
The full objective aggregates all components:
\begin{equation}\small
\label{eq:final_loss_new}
\begin{aligned}
\mathcal{L}_{\text{Critic}}
=\, &
\mathcal{L}_{\text{QR}}
+ w_{\text{risk}} \mathcal{L}_{\text{Risk}}
+ w_{\text{cvar}} \mathcal{L}_{\text{CVaR}}
+ w_{\text{gain}} \mathcal{L}_{\text{Gain}}
\\
&
+ w_{\text{shift}} \mathcal{L}_{\text{Shift}}
+ w_{\text{shape}} \mathcal{L}_{\text{Shape}}\\&
+ w_{\text{curv}} \mathcal{L}_{\text{Curv}}
+ w_{\text{consist}} \mathcal{L}_{\text{Consist}}
\end{aligned}
\end{equation}

By jointly shaping central quantiles, tail expectations, dispersion, curvature, and multi-head coherence, this calibrated objective explicitly balances robustness to noisy supervision with generalization to high-value trajectories, enabling stable and noise-resilient value distribution learning.

\begin{table*}[t]
\centering
\small
\resizebox{2\columnwidth}{!}{
\begin{tabular}{lccccc|c}
\toprule
\textbf{Domain} & \multicolumn{1}{c}{\textbf{In-Domain}} & \multicolumn{4}{c|}{\textbf{Out-of-Domain }} & \textbf{ALL}  \\
\cmidrule(lr){1-1} \cmidrule(lr){2-2} \cmidrule(lr){3-6}
\textbf{Method} & \textbf{Life Services} & \textbf{Transportation \& Travel} & \textbf{Healthcare \& Wellness} & \textbf{Social \& Entertainment} & \textbf{Financial Services} & \textbf{AVG} \\
\midrule
Baseline        & 86.73\% & 84.50\% & 90.23\% & 87.13\% & 82.70\% & 86.26\% \\
GRPO            & 30.03\% & 28.57\% & 28.33\% & 28.90\% & 27.90\% & 28.75\% \\
PPO             & 84.20\% & 86.07\% & \textbf{90.87\%}& 81.00\% & 83.87\% & 85.20\% \\
Reinforce++     & 87.87\% & 81.07\% & 77.67\% & 87.13\% & \textbf{85.53\%} & 83.85\% \\
Dr.GRPO         & 51.67\% & 60.30\% & 61.37\% & 52.17\% & 56.47\% & 56.39\% \\
\rowcolor{gray!20}\textbf{DVPO (Ours)} & \textbf{88.13\%} & \textbf{87.73\%} & 87.67\% & \textbf{87.67\%} & 82.73\% & \textbf{86.79\%} \\
\bottomrule
\end{tabular}
}
\caption{
Accuracy (\%) of different reinforcement learning methods on the \textbf{Real-world Dialog Domains} under noisy reward signals.
DVPO achieves the highest average accuracy of 86.79\%, surpassing all baselines. 
While GRPO and Dr.GRPO collapse under noise dropping below 60\%, PPO and Reinforce++ remain stable but show limited consistency across domains. 
}
\label{tab:real-world}
\end{table*}

\begin{table*}[t]
\centering
\small
\resizebox{2\columnwidth}{!}{
\begin{tabular}{lccccccc|c|c|c}
\toprule
\textbf{Domain} & \multicolumn{3}{c}{\textbf{In-Domain (Science \& QA)}} & \multicolumn{4}{c|}{\textbf{Out-of-Domain (Math)}} & \textbf{ID} & \textbf{OOD}& \textbf{ALL} \\
\cmidrule(lr){1-1} \cmidrule(lr){2-4} \cmidrule(lr){5-8}
\textbf{Method} & \textbf{SampleQA} & \textbf{GPQA(ALL)} & \textbf{HLE} & \textbf{MATH500} & \textbf{AIME24} & \textbf{Minerva-Math} & \textbf{AMC23} & \textbf{AVG} & \textbf{AVG} & \textbf{AVG}\\
\midrule
Base            & 2.89\% & 3.10\% & 2.89\% & 87.40\% & 41.67\% & 28.68\% & 75.83\% & 2.96\% & 58.40\% & 34.64\%\\

GRPO            & 3.03\% & 2.98\% & 3.24\% & 86.80\% & 45.00\% & 26.84\% & 82.50\% & 3.08\% & 60.29\%& 35.77\% \\
PPO             & 2.82\% & 2.17\% & 3.29\% & 87.00\% & 51.67\% & 29.41\% & 85.00\% & 2.76\% & 63.27\%& 37.34\% \\
Reinforce++     & 3.19\% & 4.35\% & 3.29\% & 89.40\% & 50.00\% & 30.51\% & 83.33\% & 3.61\% & 63.31\% & 37.72\%\\
Dr.GRPO         & 2.50\% & 3.99\% & 3.10\% & 88.60\% & 48.33\% & 27.21\% & 80.00\% & 3.20\% & 61.04\%& 36.25\% \\
Robust Bellman  & 3.14\% & 4.17\% & 3.38\% & 87.40\% & 45.00\% & 27.21\% & 83.33\% & 3.56\% & 60.74\%& 36.23\% \\
\rowcolor{gray!20}\textbf{DVPO (Ours)} & \textbf{3.21\%} & \textbf{4.71\%} & \textbf{3.57\%} & \textbf{90.60\%} & \textbf{56.67\%} & \textbf{31.99\%} & \textbf{86.67\%} & \textbf{3.83\%} & \textbf{66.48\%} & \textbf{39.63\%} \\
\bottomrule
\end{tabular}
}
\caption{
Accuracy (\%) of different reinforcement learning methods trained on the \textbf{Science Domain}.
DVPO achieves the highest ID average (3.83\%) and OOD average (66.48\%), showing strong robustness to noisy supervision and superior generalization across domains.
}
\label{tab:all-s}
\end{table*}

\begin{table*}[t]
\centering
\small
\resizebox{2\columnwidth}{!}{
\begin{tabular}{lccccccc|c|c|c}
\toprule
\textbf{Domain} & \multicolumn{4}{c}{\textbf{In-Domain (Math)}} & \multicolumn{3}{c|}{\textbf{Out-of-Domain (Science \& QA)}} & \textbf{ID} & \textbf{OOD} & \textbf{ALL} \\
\cmidrule(lr){1-1} \cmidrule(lr){2-5} \cmidrule(lr){6-8}
\textbf{Method} & \textbf{MATH500} & \textbf{AIME24} & \textbf{Minerva-Math} & \textbf{AMC23} & \textbf{SampleQA} & \textbf{GPQA(ALL)} & \textbf{HLE} & \textbf{AVG} & \textbf{AVG}& \textbf{AVG} \\
\midrule

Base             & 87.40\% & 41.67\% & 28.68\% & 75.83\% & 2.89\% & 3.10\% & 2.89\% & 58.40\%& 2.96\% & 34.64\%\\

GRPO            & 89.20\% & 35.00\% & 28.68\% & 84.17\% & 2.91\% & 3.99\% & 3.01\% & 59.26\% & 3.30\%& 35.28\%\\
PPO             & 86.40\% & 36.67\% & 30.51\% & 80.00\% & 2.70\% & 3.44\% & 3.34\% & 58.40\% & 3.16\%& 34.72\%\\
Reinforce++     & 89.00\% & 45.00\% & 27.21\% & 84.17\% & 3.19\% & 4.35\% & \textbf{3.61\%} & 61.35\% & 3.72\%& 36.65\% \\
Dr.GRPO         & 87.80\% & 45.00\% & 27.94\% & 85.83\% & 3.10\% & 4.35\% & 3.20\% & 61.64\% & 3.55\%& 36.75\% \\

Robust Bellman  & 89.40\% & 36.67\% & \textbf{31.99\%} & 84.17\% & 2.61\% & 3.62\% & 3.43\% & 60.56\% & 3.22\%& 35.98\% \\
\rowcolor{gray!20}\textbf{DVPO (Ours)} & \textbf{90.00\%} & \textbf{56.67\%} & 31.62\% & \textbf{87.50\%} & \textbf{3.31\%} & \textbf{5.25\%} & 3.57\% & \textbf{66.45\%} & \textbf{4.04\%} & \textbf{39.70\%}\\
\bottomrule
\end{tabular}
}
\caption{
Accuracy (\%) of different reinforcement learning methods trained on the \textbf{Math Domain}.
DVPO significantly outperforms all baselines, achieving an ID average of 66.45\% and OOD average of 4.04\%, indicating better generalization across noisy tasks.
}
\label{tab:all-m}
\end{table*}

\begin{table*}[t]
\centering
\small
\resizebox{2\columnwidth}{!}{
\begin{tabular}{lccccccc|c|c|c}
\toprule
\textbf{Domain} & \multicolumn{3}{c}{\textbf{In-Domain (Science \& QA)}} & \multicolumn{4}{c|}{\textbf{Out-of-Domain (Math)}} & \textbf{ID} & \textbf{OOD}& \textbf{ALL} \\
\cmidrule(lr){1-1} \cmidrule(lr){2-4} \cmidrule(lr){5-8}   
\textbf{Risk Interval} & \textbf{SampleQA} & \textbf{GPQA(ALL)} & \textbf{HLE} & \textbf{MATH500} & \textbf{AIME24} & \textbf{Minerva-Math} & \textbf{AMC23} & \textbf{AVG} & \textbf{AVG} & \textbf{AVG} \\
\midrule
Base            & 2.89\% & 3.10\% & 2.89\% & 87.40\% & 41.67\% & 28.68\% & 75.83\% & 2.96\% & 58.40\% & 34.64\%\\
0 & 3.05\% & \textbf{5.25\%} & 3.52\% & 90.40\% & 45.00\% & 28.68\% & 79.17\% & \textbf{3.94\%} & 60.81\% & 36.44\% \\
0.05 & 3.10\% & 4.89\% & 3.52\% & 90.20\% & 46.67\% & 31.25\% & 79.17\% & 3.84\% & 61.82\% & 36.97\% \\
\rowcolor{gray!20}\textbf{0.1 (Ours)} & 3.21\% & 4.71\% & \textbf{3.57\%} & \textbf{90.60\%} & \textbf{56.67\%} & \textbf{31.99\%} & \textbf{86.67\%} & 3.83\% & \textbf{66.48\%} & \textbf{39.63\%} \\
0.2 & \textbf{3.28\%} & 4.35\% & 3.15\% & 90.00\% & 46.67\% & 29.41\% & 85.00\% & 3.59\% & 62.77\% & 37.41\% \\
\bottomrule
\end{tabular}
}
\caption{
Comparison of DVPO performance under different risk interval settings across multiple domains. 
A moderate interval weight of 0.1 provides the most balanced and stable results, while both smaller and larger intervals degrade robustness and generalization. 
}
\label{tab:ablation-fx}
\end{table*}

\begin{table*}[t]
\centering
\small
\resizebox{2\columnwidth}{!}{
\begin{tabular}{lccccccc|c|c|c}
\toprule
\textbf{Domain} & \multicolumn{3}{c}{\textbf{In-Domain (Science \& QA)}} & \multicolumn{4}{c|}{\textbf{Out-of-Domain (Math)}} & \textbf{ID} & \textbf{OOD}& \textbf{ALL} \\
\cmidrule(lr){1-1} \cmidrule(lr){2-4} \cmidrule(lr){5-8}
\textbf{Interval Density} & \textbf{SampleQA} & \textbf{GPQA(ALL)} & \textbf{HLE} & \textbf{MATH500} & \textbf{AIME24} & \textbf{Minerva-Math} & \textbf{AMC23} & \textbf{AVG} & \textbf{AVG} & \textbf{AVG} \\
\midrule
Base            & 2.89\% & 3.10\% & 2.89\% & 87.40\% & 41.67\% & 28.68\% & 75.83\% & 2.96\% & 58.40\% & 34.64\%\\
50          & 2.84\% & 3.62\% & 3.06\% & 87.60\% & 46.67\% & \textbf{31.99\%} & 83.33\% & 3.17\% & 62.40\% & 37.02\% \\
100         & 2.94\% & 3.99\% & 3.20\% & 88.20\% & 55.00\% & \textbf{31.99\%} & 83.33\% & 3.38\% & 64.63\% & 38.38\% \\
\rowcolor{gray!20}\textbf{200 (Ours)} & \textbf{3.21\%} & \textbf{4.71\%} & \textbf{3.57\%} & \textbf{90.60\%} & \textbf{56.67\%} & \textbf{31.99\%} & \textbf{86.67\%} & \textbf{3.83\%} & \textbf{66.48\%} & \textbf{39.63\%} \\
500          & 2.84\% & 3.80\% & 3.34\% & 89.40\% & 46.67\% & 31.62\% & 85.83\% & 3.33\% & 63.38\% & 37.64\% \\
\bottomrule
\end{tabular}
}
\caption{
Comparison of DVPO results under different interval densities across multiple domains. 
A moderate density of 200 yields the most stable and accurate performance, while both overly sparse and overly dense settings reduce robustness and generalization under noisy supervision. 
}
\label{tab:ablation-interval-density}
\end{table*}

\section{Experiments}
\subsection{Setup}
We evaluate DVPO in math, science and dialogue tasks under rule-based rewards and model-based rewards supervision.
\paragraph{Dialogue Task} To assess the effectiveness of our approach in realistic settings with model-based reward supervision, we conduct multi-turn dialogue experiments on the \textbf{Honor-Dialogue Dataset}. This dataset, constructed by us, contains multi-domain, task-oriented dialogues collected from real-world scenarios. The task requires the model to act as a dialogue assistant, producing natural and effective responses. Under noisy dialogue reward model supervision, the model is trained with reinforcement learning on real multi-turn dialogues. Performance is evaluated through interactions with GPT-4o using the dialogue standards in Section~4.2. Given that the performance of some methods degrades drastically as training steps increase, we use models trained for limited steps for experimental comparison. All real information in the data has undergone anonymization, and the use of the data has been authorized and reviewed by the ethics committee.

\paragraph{Math and Science Task} To further assess the effectiveness of our approach in realistic settings where supervision comes from rule-based rewards, we conduct a series of experiments on mathematical and scientific reasoning tasks during training. Specifically, we use the Light-R1 dataset \cite{wen2025lightr1curriculumsftdpo} for math tasks and the SuperGPQA\cite{pteam2025supergpqascalingllmevaluation} dataset for scientific tasks. For these datasets, we first conduct multiple inference runs with the original model and then adopt majority voting to filter the results as training labels prior to model training. The evaluation is then conducted on math and scientific QA datasets. The test datasets include four math-focused tasks: MATH500 \cite{hendrycks2021measuringmathematicalproblemsolving}, AIME24, Minerva-Math \cite{lewkowycz2022solvingquantitativereasoningproblems}, AMC23, as well as three scientific knowledge tasks: SampleQA \cite{wei2024measuringshortformfactualitylarge}, GPQA , and HLE (Humanity’s Last Exam)\cite{phan2025humanitysexam}. Table~\ref{tab:all-s} reports the accuracy of models before and after RL training. Due to the limitation of memory, multiple rounds of generation are carried out here. In each round, the generated length is at most 4096 tokens, and the result of the last round is used for evaluation.

\begin{figure}[h]
\centering
\includegraphics[width=1 \linewidth]{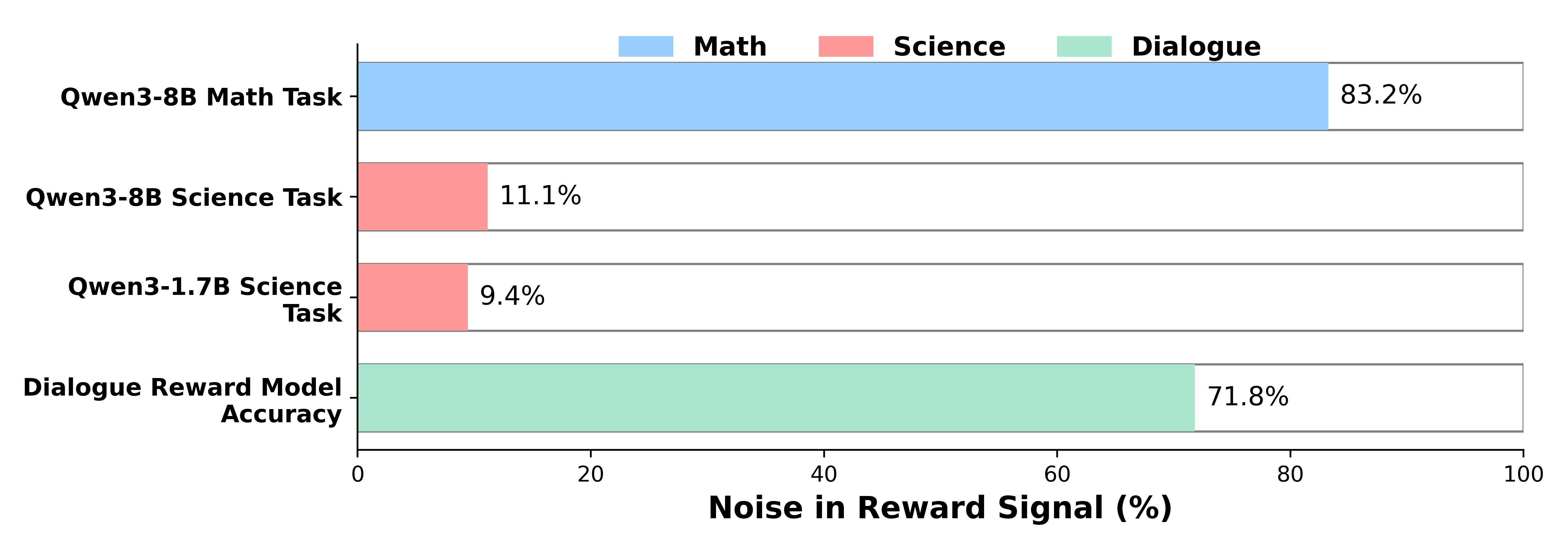}
\caption{Noise statistics in the various tasks. A significant portion of rewards contains inaccuracies.}
\label{fig:task-noise}
\end{figure}

\paragraph{Baseline Initialization}In the math and scientific tasks, we use Qwen3-8B\cite{qwen3} for initialization. In the dialogue tasks, both the policy and reward models are initialized from Qwen3-8B and then fine-tuned on the Honor-Dialogue dataset.
In our experiments, we implemented a PPO method based on the robust Bellman operator. Specifically, a multi-head value model is constructed to assign multiple value estimates to each token, which are then aggregated by selecting the minimum value as the current estimate to simulate the worst-case scenario. This worst-case value serves as the optimization target, guiding the overall value model training toward robustness under adverse conditions.
\paragraph{Training Noise} Noise is pervasive in RL training , especially under model-based feedback. Figure~\ref{fig:task-noise} illustrates the error rate in various reward signals. The noise in the rule-based reward from the majority vote generated data.

\paragraph{Implementation Details} A majority vote of 5 samples is conducted. The risk threshold is set at 0.1, the number of interval quantile points is set to 200 and the number of distributional value head is set to 3. During training, the model generates dialogue tasks of length 4096 for reasoning tasks and 1024 for dialogue tasks. The experiments are conducted on 8 * NVIDIA A100 80GB GPUs.For further dataset construction, training details, and model initialization, please refer to Appendix.

\subsection{Evaluation Metrics}
 For mathematical reasoning and scientific knowledge tasks, we adopt accuracy as the primary metric to measure whether the model produces correct answers. For multi-turn dialogue tasks, we employ a tripartite evaluation framework that captures both task execution and communicative quality: (1) Task Completion Rate (TCR), assessing whether the model successfully fulfills the user-intended objective; (2) Ask Completion Rate (ACR), measuring whether detailed aspects of the user query are adequately addressed; and (3) Goal Completion Rate (GCR), evaluating the fluency, coherence, and appropriateness of the model's responses. This comprehensive evaluation allows us to examine not only whether the model achieves the intended outcomes, but also how effectively it aligns with human communication standards.



\subsection{Experiment results}


\paragraph{DVPO exhibits strong robustness under real noisy supervision}
As shown in Table~\ref{tab:real-world}, Table~\ref{tab:all-s}, and Table~\ref{tab:all-m}, DVPO consistently improves performance across diverse domains even with imperfect reward signals. 
In dialogue tasks, the average task completion rate increases from 71.48\% to 73.66\%, while both GRPO and Dr.GRPO exhibit severe performance degradation under noisy rewards, reflecting the effectiveness of DVPO’s distributional value supervision. 
In mathematical and scientific reasoning tasks, DVPO maintains strong performance, achieving average accuracies of 39.70\% and 39.63\% respectively, substantially surpassing PPO and GRPO, demonstrating its robustness in multi-domain noisy training.


\paragraph{DVPO achieves reliable balance in in-domain and cross-domain performance}
As shown in Table~\ref{tab:real-world}, Table~\ref{tab:all-s}, and Table~\ref{tab:all-m}, DVPO demonstrates consistent advantages across dialogue, mathematical, and scientific domains under noisy reward supervision. 
In real-world dialogue tasks, DVPO achieves an average accuracy of 88.13\% in the life services domain, while maintaining strong performance across other domains, outperforming all baseline methods. 
When trained on mathematical or scientific reasoning tasks, DVPO attains the highest in-domain accuracy and exhibits remarkable out-of-domain generalization, consistently surpassing PPO and robust Bellman-based approaches. 
Notably, while Robust Bellman PPO exhibits stable training loss and an optimistic reward trajectory during training, it suffers from poor generalization outside the training domain. It frequently generates lengthy, incoherent responses on unseen tasks and even fails to maintain the required response format, which makes it impossible to present its results in the dialogue task.
In contrast, DVPO achieves a good balance between stability and generalization by combining distributional value modeling with conditional risk control, showing strong performance both within and across domains.

\subsection{Ablation Experiment and Discussion}
To further analyze the impact of the DVPO method, an experimental ablation study is conducted on the scientific task. The SuperGPQA dataset is still used for training, and the parameters remain the same as those in the main experiment.

\paragraph{A proper interval density helps the model learn more effectively from value distributions}
As shown in Table~\ref{tab:ablation-interval-density}, when the interval density decreases to 50, the average performance drops from 39.63\% to 38.38\% and further to 37.02\%. 
When the density becomes excessively high, up to 500, performance also declines to 37.64\%, with both in-domain and out-of-domain results deteriorating noticeably. 
During the early phase of training, using a slightly higher interval density helps the model receive richer supervision signals, giving more consistent feedback under noisy rewards and improving training stability. 
However, when the intervals become too fine-grained, the model tends to capture noise as useful information, which reduces robustness. 
In contrast, moderately sparse intervals allow the model to focus on key informative regions while keeping the distribution smoother, leading to better generalization and stability.

\paragraph{A proper level of risk interval control enhances robustness and generalization}
As shown in Table~\ref{tab:ablation-fx}, removing risk interval control entirely (weight = 0) increases local accuracy on GPQA but leads to poor generalization and unstable training across mathematical domains. 
When the interval weight becomes too large, excessive perturbations in the value distribution distort the relationship between training and generation distributions, causing instability and lower overall performance. 
In contrast, DVPO with a moderate interval setting 0.1 achieves the best trade-off. It suppresses noise effectively while maintaining adequate exploration, resulting in the highest average accuracy and stable optimization behavior across domains.

\section{Conclusion}

We introduce DVPO, a robust RL framework that integrates distributional value modeling with conditional risk control in LLM post-training. DVPO offers a novel perspective on balancing robustness and generalization, achieving a principled trade-off between the two under noisy supervision by modeling value distributions and regulating their risk-aware tail structure. Experiments across dialogue, mathematical, and scientific tasks verify its superiority over PPO, GRPO, and robust Bellman PPO methods, providing a scalable and effective solution for real-world RL.


\section*{Limitations}

While DVPO demonstrates strong robustness and generalization under noisy supervision, several limitations remain. The optimal choice of distributional interval density and risk thresholds may vary across domains, requiring task-specific tuning for best performance. And, although DVPO improves stability under moderate noise, extreme reward corruption or mis-specified supervision can still degrade performance.

\bibliography{custom}

\appendix

\section{Additional Details for VRPO}
\label{sec:appendix}
\subsection{Pseudocode}
The full algorithm of VRPO is detailed in Algorithm ~\ref{alg:dvpo}.

\begin{algorithm}
\caption{DVPO Training Algorithm}
\label{alg:dvpo}
\begin{algorithmic}[1]
\small
\REQUIRE Dataset $\mathcal{D}$, policy $\pi_\theta$, value network $\{f_{\phi_i}\}_{i=1}^N$
\ENSURE Optimized $\pi_\theta$ and $\{f_{\phi_i}\}$

\FOR{each iteration}
    \STATE \textbf{Sample trajectories} using $\pi_\theta$
    
    \STATE \textbf{Distributional Value Estimation:}
    \FOR{each $(s_t, a_t)$}
        \STATE $h_t = \text{Encoder}(s_t)$
        \STATE $\theta_j^{\text{ens}} = \frac{1}{N}\sum_{i=1}^N f_{\phi_i}(h_t)$
    \ENDFOR
    
    \STATE \textbf{Compute Distributional GAE:}
    \STATE $\Theta_{A_t} = (r_t + \gamma\Theta_{V_{t+1}} - \Theta_{V_t}) + \gamma\lambda\Theta_{A_{t+1}}$
    \STATE $A_t = \frac{1}{M}\sum_j \Theta_{A_t,j}$
    
    \STATE \textbf{Value Model Update:}
    \STATE $\mathcal{L}_{\text{QR}} = \mathbb{E}[\frac{1}{M}\sum_j |\tau_j - I(u_j<0)| L_\delta(u_j)]$
    
    \STATE $\mathcal{L}_{\text{Risk}} = \mathbb{E}[\frac{1}{M}\sum_j w_{\text{risk}}(\tau_j)$
    \STATE \quad $|\tau_j - I(u_j<0)| L_\delta(u_j)]$
    
    \STATE $\mathcal{L}_{\text{CVaR}} = \mathbb{E}[(\frac{1}{K_\alpha}\sum_{j=1}^{K_\alpha} \theta_{\pi(j)}' - \theta_{\pi(j)})^2]$
    
    \STATE $\mathcal{L}_{\text{Gain}} = \mathbb{E}[(\frac{1}{K_\beta}\sum_{j=M-K_\beta+1}^M \theta_{\pi(j)}' - \theta_{\pi(j)})^2]$
    
    \STATE $\mathcal{L}_{\text{Shift}} = \mathbb{E}[\text{ReLU}(\mathbb{E}[\theta_j'] - \mathbb{E}[\theta_j])]$
    
    \STATE $\mathcal{L}_{\text{Shape}} = \mathbb{E}[\text{ReLU}(\text{Var}(\Theta;\mathcal{I}_\alpha) - \text{Var}(\Theta';\mathcal{I}_\alpha))$ 
    \STATE \quad $+ \text{ReLU}(\text{Var}(\Theta';\mathcal{I}_\beta) - \text{Var}(\Theta;\mathcal{I}_\beta))]$
    
    \STATE $\mathcal{L}_{\text{Curv}} = \mathbb{E}[\text{ReLU}(\mathbb{E}_{j\in\mathcal{I}_\alpha}[\Delta^2\theta_j])$
    \STATE \quad $+ \text{ReLU}(-\mathbb{E}_{j\in\mathcal{I}_\beta}[\Delta^2\theta_j])]$
    
    \STATE $\mathcal{L}_{\text{Consist}} = \frac{1}{N_{\text{pair}}}\sum_{i<j} \mathbb{E}[\|\Theta^{(i)} - \Theta^{(j)}\|_2^2]$
    
    \STATE $\mathcal{L}_{\text{Critic}} = \sum_k w_k \mathcal{L}_k$
    \STATE Update $\{\phi_i\}$ with $\nabla\mathcal{L}_{\text{Critic}}$
    
    \STATE \textbf{Policy Update:}
    \STATE $\mathcal{L}_{\text{PPO}} = \mathbb{E}[\min(\rho_t A_t, \text{clip}(\rho_t,1-\epsilon,1+\epsilon)A_t)]$
    \STATE Update $\theta$ with $\nabla\mathcal{L}_{\text{PPO}}$
\ENDFOR
\end{algorithmic}
\end{algorithm}

\section{Mathematical Analysis of Stability and Generalization in  Robust Bellman PPO and DVPO}

This section provides a rigorous mathematical derivation of the training dynamics for Standard PPO, Robust Bellman PPO, and our DVPO framework. We demonstrate why the pessimistic contraction of Robust Bellman PPO leads to stability at the cost of generalization, and how DVPO's asymmetric objective, validated by Figure ~\ref{fig:adv-fenbu}, achieves a principled balance.

\subsection{Instability of Standard PPO and Symmetric Contraction of Robust Bellman PPO}

The critic $V_\phi(s)$ in a standard PPO framework is trained to minimize the Mean Squared Error loss with respect to a return target $Y_t$, $\mathcal{L}_{Critic}^{PPO}(\phi) = E_t [ (Y_t - V_\phi(s_t))^2 ]$. When supervision is noisy, the target $Y_t$ (e.g., $r_t + \gamma V_\phi(s_{t+1})$) becomes a high-variance random variable. This high target variance, $Var(Y_t)$, induces high-variance gradients $\nabla_\phi \mathcal{L}_{Critic}^{PPO}$, destabilizing the optimization process.

A Robust Bellman PPO  baseline, as described in our implementation and motivated by robust MDP theory, adopts a pessimistic formulation to ensure stability. It defines the robust value estimate as the minimum over all heads, $V_{robust}(s) = \min_i V_{\phi, i}(s)$. The $\min$ operator functions as a non-linear filter that is inherently robust to upper-tail noise (i.e., anomalously high reward signals). This operation dramatically reduces the target variance, $Var(Y_t^{Robust}) \ll Var(Y_t^{PPO})$, yielding low-variance gradients and a stable training loss.

However, this stability comes at a high cost. The $\min$ operator is a symmetric contractor in effect, it discards all information not related to the worst-case outcome. This mathematical collapse is empirically verified by Figure ~\ref{fig:adv-fenbu} in the main text. As shown, the Robust Bellman PPO value distribution (red) collapses toward its lower bound and is concentrated with small variance. By discarding all optimistic signals, this method mathematically precludes the policy from learning about high-potential opportunities, thereby severely degrading generalization.

\subsection{Asymmetric Regulation for Balanced Optimization}

DVPO resolves this dilemma through its Conditional Risk-aware Tail-Constrained (CRTC) objective. This mechanism is mathematically grounded in the ReLU-based tail shape losses, which create one-way optimization penalties to regulate the tails asymmetrically.

\subsubsection{Robustness via Lower-Tail Contraction}
Robustness against pessimistic noise is achieved via the lower-tail shape loss:
\begin{equation} \small
\mathcal{L}_{Shape}^{lower} = E_{s,a}[ReLU(Var(\Theta;\mathcal{I}_{\alpha}) - Var(\Theta^{\prime};\mathcal{I}_{\alpha}))]
\end{equation}
Let $V_{pred}^{lower} = Var(\Theta;\mathcal{I}_{\alpha})$ be the predicted lower-tail variance and $V_{target}^{lower} = Var(\Theta^{\prime};\mathcal{I}_{\alpha})$ be the target. The loss is $\mathcal{L} = E[ReLU(V_{pred}^{lower} - V_{target}^{lower})]$. The gradient with respect to the predicted variance, $\frac{\partial \mathcal{L}}{\partial V_{pred}^{lower}}$, is key. If $V_{pred}^{lower} > V_{target}^{lower}$, the ReLU is active, yielding a gradient of $+1$. Gradient descent applies a penalty to decrease $V_{pred}^{lower}$. Conversely, if $V_{pred}^{lower} \le V_{target}^{lower}$, the gradient is $0$, and no penalty is applied. This creates a one-way upper bound, enforcing $V_{pred}^{lower} \le V_{target}^{lower}$. This mathematical contraction of the lower tail filters pessimistic noise. This exact outcome is confirmed by Figure 3, which shows the DVPO distribution (green) also contracts slightly in the lower tail, achieving the desired robustness.

\subsubsection{Generalization via Upper-Tail Expansion}
Generalization is preserved via the upper-tail shape loss, which inverts the arguments:
\begin{equation} \small
\mathcal{L}_{Shape}^{upper} = E_{s,a}[ReLU(Var(\Theta^{\prime};\mathcal{I}_{\beta}) - Var(\Theta;\mathcal{I}_{\beta}))]
\end{equation}
Let $V_{pred}^{upper} = Var(\Theta;\mathcal{I}_{\beta})$ and $V_{target}^{upper} = Var(\Theta^{\prime};\mathcal{I}_{\beta})$. The loss is $\mathcal{L} = E[ReLU(V_{target}^{upper} - V_{pred}^{upper})]$. The gradient $\frac{\partial \mathcal{L}}{\partial V_{pred}^{upper}}$ is asymmetric in the opposite direction. If $V_{pred}^{upper} < V_{target}^{upper}$, the ReLU is active, yielding a gradient of $-1$. Gradient descent (moving in the negative gradient direction, i.e., $-(-1) = +1$) will increase $V_{pred}^{upper}$ toward the target. Conversely, if $V_{pred}^{upper} \ge V_{target}^{upper}$, the gradient is $0$, and no penalty is applied. This creates a one-way lower bound, enforcing $V_{pred}^{upper} \ge V_{target}^{upper}$.

This upper-tail expansion preserves the variance of high-value outcomes, preventing the policy from becoming overly conservative. Figure ~\ref{fig:adv-fenbu} provides direct empirical validation of this mathematical mechanism. In stark contrast to Robust Bellman PPO, the DVPO distribution (green) achieves significant exploratory expansion learning in the upper tail. This visual evidence confirms that our asymmetric ReLU loss successfully translates from mathematical theory to practical effect, achieving the principled balance of robustness and generalization that Robust Bellman PPO's symmetric contraction fails to capture.

\section{Additional Experimental Details}
\label{appendix:setup}
\subsection{Setup}
\subsubsection{Dataset Construction}
\paragraph{Math and scientific tasks}
For mathematical tasks, taking the Qwen3-8B model as an illustrative example, pseudo-labels are generated from 39,000 samples in the Light-R1 dataset \cite{wen2025lightr1curriculumsftdpo}. Following 5 rounds of majority voting, samples with at least 3 identical responses are retained, resulting in 31,209 instances for reinforcement learning (RL) training.
For scientific tasks, pseudo-labels are generated from 26,529 samples in the SuperGPQA dataset. After the same 5-round majority voting process, 10,075 samples with at least 2 consistent responses are preserved for RL training.
To ensure fairness, evaluation is performed on mathematical benchmarks that do not overlap with the training datasets.

\paragraph{Real-dialogue tasks}
A real-world dialogue dataset Honor-Dialogue is used.    The reward model is trained on 36,000 labeled samples, with 3,000 reserved for validation.    The policy model is fine-tuned and subsequently trained via reinforcement learning on an additional 50,952 conversations, excluding data from the financial services, social entertainment, transportation, and healthcare domains.
Performance evaluation is conducted across five scenarios: 1,004 conversations for daily services, 700 for financial services, 534 for social entertainment, 463 for transportation, and 380 for healthcare.
This dataset encompasses diverse dialogue tasks in real-life scenarios, accurately reflecting practical real-world conditions.    Notably, \textbf{such multi-category real-scenario dialogue task data is not available in other datasets.}

Honor-Dialogue dataset is constructed based on a goal-driven scenario-oriented design paradigm.    Specifically, we first defined the core role of the AI dialogue agent, a practical, scenario-adaptive assistant, and established strict constraints on response accuracy, naturalness, and compliance with real-world service norms.    Subsequently, we selected six representative domains including one general service domain and five specific professional domains.    These domains are derived from high-frequency real-life interaction scenarios, each equipped with explicit trigger conditions such as user query intents and scenario context features, as well as standardized response rules such as information provision priority and dialogue flow guidelines.

For the data within each domain, we construct realistic caller inputs that include scenario-matched latest messages and conversation history, as well as corresponding standardized outputs that comply with the requirements of the target goal.   We explicitly mark the dialogue state, response content, and matched target to ensure the quality of the supervised training data.

A representative example of the Honor-Dialogue dataset is presented below.(Figure ~\ref{fig:dialogue-example})    The notation xxx denotes the masking of sensitive information such as user ID and contact details, which is implemented to comply with data privacy regulations.
\begin{table*}[t]
\centering
\small
\resizebox{2\columnwidth}{!}{
\begin{tabular}{lccccccccccc|c}
\toprule
\textbf{Scenario} & Wealth & Rental & Insurance & Food & Express & Promotion & Loan & Housing & Service & Product & General & Avg \\
\midrule
Dialogue Count & 87 & 99 & 138 & 120 & 215 & 66 & 70 & 87 & 67 & 69 & 92 & 94.73 \\
Avg Turns      & 5.40 & 4.22 & 5.22 & 3.37 & 3.76 & 4.55 & 4.33 & 4.68 & 5.46 & 4.44 & 3.46 & 4.44 \\
\bottomrule
\end{tabular}
}
\caption{
Dataset statistics across 11 real-world service scenarios.
The validation set covers diverse interaction types with varying dialogue lengths, enabling reliable evaluation of robustness and generalization under realistic settings.
}
\label{tab:validation-data-statistics}
\end{table*}
\begin{table*}[ht]
\centering
\small
\resizebox{0.7\linewidth}{!}{
\begin{tabular}{lcccccc|cc}
\toprule
\textbf{Category} & \multicolumn{4}{c}{\textbf{Human Annotators}} & \multicolumn{2}{c|}{\textbf{GPT Models}} & \textbf{Human AVG} & \textbf{Model AVG}  \\
\cmidrule(lr){1-1} \cmidrule(lr){2-5} \cmidrule(lr){6-7} \cmidrule(lr){8-9}
Score & 4.595  & 4.689  & 4.490  & 4.330  & 4.510  & 4.501  & 4.526     & 4.505     \\
\bottomrule
\end{tabular}
}
\caption{Comparison of goal completion rate scores between human annotators and GPT-4o. 
Human annotators achieve an average score of 4.526, while GPT-4o models reach 4.505, showing a negligible performance gap.}
\label{tab:human-model-agreement}
\end{table*}

\begin{table*}[t]
\centering
\small
\resizebox{2\columnwidth}{!}{
\begin{tabular}{lcccc|cccc|cccc|cccc|cccc|c}
\toprule
\multirow{2}{*}{\textbf{Method}} &
\multicolumn{4}{c|}{\textbf{Life Services (In-Domain)}} &
\multicolumn{4}{c|}{\textbf{Transportation \& Travel}} &
\multicolumn{4}{c|}{\textbf{Healthcare \& Wellness}} &
\multicolumn{4}{c|}{\textbf{Social \& Entertainment}} &
\multicolumn{4}{c|}{\textbf{Financial Services}} &
\multirow{2}{*}{\textbf{AVG}} \\
\cmidrule(lr){2-21}
& TC & AC & GC & D-AVG
& TC & AC & GC & D-AVG
& TC & AC & GC & D-AVG
& TC & AC & GC & D-AVG
& TC & AC & GC & D-AVG
& \\
\midrule
Baseline 
& 72.5 & 95.0 & 92.7 & 86.73
& 66.7 & 96.5 & 90.3 & 84.50
& 77.2 & \textbf{97.8} & 95.7 & 90.23
& 75.3 & 92.0 & \textbf{94.1} & 87.13
& 65.7 & 91.4 & 91.0 & 82.70
& 86.26
\\
GRPO
& 42.7 & 16.3 & 31.1 & 30.03
& 34.6 & 20.6 & 30.5 & 28.57
& 33.3 & 19.4 & 32.3 & 28.33
& 46.4 & 13.1 & 27.2 & 28.90
& 49.1 & 13.3 & 21.3 & 27.90
& 28.75
\\
PPO
& 69.0 & 93.1 & 90.5 & 84.20
& 71.4 & 96.0 & 90.8 & 86.07
& \textbf{82.5} & 93.1 & \textbf{97.0} & \textbf{90.87}
& 67.0 & 87.9 & 88.1 & 81.00
& \textbf{69.2} & 96.3 & 86.1 & 83.87
& 85.20
\\
Reinforce++
& 73.9 & \textbf{96.5} & \textbf{93.2} & 87.87
& 58.7 & \textbf{96.8} & 87.7 & 81.07
& 47.9 & 95.3 & 89.8 & 77.67
& 70.3 & \textbf{98.6} & 92.5 & 87.13
& 68.8 & \textbf{96.5} & \textbf{91.3} & \textbf{85.53}
& 83.85
\\
Dr.GRPO
& 37.1 & 51.8 & 66.1 & 51.67
& 39.7 & 68.0 & 73.2 & 60.30
& 33.3 & 75.4 & 75.4 & 61.37
& 40.6 & 52.6 & 63.3 & 52.17
& 39.6 & 63.7 & 66.1 & 56.47
& 56.39
\\
\rowcolor{gray!20}
\textbf{DVPO (Ours)}
& \textbf{76.5} & 95.7 & 92.2 & \textbf{88.13}
& \textbf{75.6} & 95.8 & \textbf{91.8} & \textbf{87.73}
& 73.2 & 96.9 & 92.9 & 87.67
& \textbf{76.3} & 94.5 & 92.2 & \textbf{87.67}
& 66.7 & 90.7 & 90.8 & 82.73
& \textbf{86.79}
\\
\bottomrule
\end{tabular}
}
\caption{
Full results on the Real-world Dialogue Benchmark under noisy reward supervision. 
For each domain, we report task completion rate (TC), ask completion rate (AC), and goal completion rate (GC); DomainAVG (D-AVG) denotes the averaged performance across the three metrics within each domain, and AVG denotes the overall average across all five domains. 
DVPO achieves the highest overall average performance 86.79\%, consistently surpassing all prior RL baselines in both in-domain and out-of-domain settings.
}
\label{tab:dialog-full}
\end{table*}

\begin{table*}[t]
\centering
\small
\resizebox{2\columnwidth}{!}{
\begin{tabular}{lccccccc|c|c|c}
\toprule
\textbf{Domain} & \multicolumn{3}{c}{\textbf{In-Domain (Science \& QA)}} & \multicolumn{4}{c|}{\textbf{Out-of-Domain (Math)}} & \textbf{ID} & \textbf{OOD}& \textbf{ALL} \\
\cmidrule(lr){1-1} \cmidrule(lr){2-4} \cmidrule(lr){5-8}  
\textbf{Method} & \textbf{SampleQA} & \textbf{GPQA(ALL)} & \textbf{HLE} & \textbf{MATH500} & \textbf{AIME24} & \textbf{Minerva-Math} & \textbf{AMC23} & \textbf{AVG} & \textbf{AVG} & \textbf{AVG} \\
\midrule

Base & 1.78 & 2.36 & 2.87 & 84.80 & 35.00 & 19.49 & 69.17 & 2.34 & 52.12 & 30.78 \\
GRPO & 1.60 & 1.99 & 3.94 & 82.20 & 33.33 & 21.32 & 77.50 & 2.51 & 53.59 & 31.70 \\
PPO & \textbf{1.91} & \textbf{2.54} & 3.94 & 83.80 & 33.33 & 22.43 & \textbf{79.17} & 2.80 & 54.68 & 32.45 \\
Reinforce++ & 1.48 & 2.36 & 3.52 & 84.60 & 38.33 & 23.16 & 75.83 & 2.45 & 55.48 & 32.75 \\
Dr.GRPO & 1.50 & 2.17 & 3.10 & 84.00 & 35.00 & 22.79 & 76.67 & 2.26 & 54.62 & 32.18 \\
Robust Bellman & 1.64 & 2.17 & 4.17 & 84.60 & 31.67 & 22.79 & 75.00 & 2.66 & 53.52 & 31.72 \\
\rowcolor{gray!20}\textbf{DVPO (Ours)} & 1.73 & 2.36 & \textbf{4.36} & \textbf{86.20} & \textbf{43.33} & \textbf{25.00} & 78.33 & \textbf{2.82} & \textbf{58.22} & \textbf{34.47} \\
\bottomrule
\end{tabular}
}
\caption{
Comparison of DVPO performance on the smaller Qwen3-1.7B model across multiple domains.  DVPO achieved the best performance in math tasks, and also achieved the optimal accuracy rate of 4.36\% on the HLE dataset.
}
\label{tab:small-model}
\end{table*}

\begin{table*}[t]
\centering
\small
\resizebox{2\columnwidth}{!}{
\begin{tabular}{lccccccc|c|c|c}
\toprule
\textbf{Domain} & \multicolumn{3}{c}{\textbf{In-Domain (Science \& QA)}} & \multicolumn{4}{c|}{\textbf{Out-of-Domain (Math)}} & \textbf{ID} & \textbf{OOD}& \textbf{ALL} \\
\cmidrule(lr){1-1} \cmidrule(lr){2-4} \cmidrule(lr){5-8}
\textbf{Method} &
\textbf{SampleQA} &
\textbf{GPQA (ALL)} &
\textbf{HLE} &
\textbf{MATH500} &
\textbf{AIME24} &
\textbf{Minerva-Math} &
\textbf{AMC23} &
\textbf{AVG} &
\textbf{AVG} &
\textbf{AVG} \\
\midrule
Base            & 2.89\% & 3.10\% & 2.89\% & 87.40\% & 41.67\% & 28.68\% & 75.83\% & 2.96\% & 58.40\% & 34.64\%\\
Core Quantile Regression & 2.94\% & 4.17\% & \textbf{3.85\%} & 90.80\% & 45.00\% & 30.51\% & 79.17\% & 3.65\% & 61.37\% & 36.63\% \\
Above+Distribution Consistency & 2.87\% & 4.53\% & 3.80\% & 90.40\% & 51.67\% & 28.68\% & 84.17\% & 3.73\% & 63.73\% & 38.02\% \\
Above+Tail Calibration & 2.77\% & \textbf{4.89\%} & 3.57\% & 89.40\% & 51.67\% & 31.62\% & 85.00\% & 3.74\% & 64.42\% & 38.42\% \\
Above+Shift Penalization & 2.94\% & 3.44\% & 3.48\% & \textbf{91.20\%} & 51.67\% & \textbf{31.99\%} & 85.83\% & 3.29\% & 65.17\% & 38.65\% \\
\rowcolor{gray!20}\textbf{Above+Tail Shape \& Curvature(OURS)} & \textbf{3.21\%} & 4.71\% & 3.57\% & 90.60\% & \textbf{56.67\%} & \textbf{31.99\%} & \textbf{86.67\%} & \textbf{3.83\%} & \textbf{66.48\%} & \textbf{39.63\%} \\
\bottomrule
\end{tabular}
}
\caption{
Ablation study on the effect of different loss components in DVPO.  
Each loss is added on the above experiment. 
All components provide positive contributions, and performance improves steadily as more distributional constraints are incorporated.  
Tail-focused objectives yield the most significant gains, Tail Calibration improves OOD accuracy from 61.37\% to 64.42\%, while adding Tail Shape and Curvature further enhances both ID and OOD performance, achieving the best overall result of 39.63\%.  
}
\label{tab:dvpo-ablation}
\end{table*}

\begin{figure*}[!t]
\centering

\begin{tcolorbox}[
  colback=gray!4!white,
  colframe=gray!85!black,
  title=\textbf{System Prompt},
  fonttitle=\bfseries
]
\textbf{Role}: You are the AI call assistant of xxx Inc., capable of answering calls clearly and politely on behalf of the user and conversing with the caller. Based on the user's customized dialogue goals, you should first determine which goal applies to the conversation and respond accordingly. You do not know any personal information about the user beyond the task description and must inform the caller that any information they provide will be passed on to the user.

\textbf{Dialogue Goals (Customized)}:
\begin{itemize}
  \item \textbf{0: General Domain} – Used if the call doesn't match any specific domain. Be polite and professional. Try to answer or guide the caller and clarify their intent.
  \item \textbf{1: E-commerce/Retail} – If caller mentions down jacket performance, mention fast delivery, good warmth at -5°C, ask for cleaning tips, other styles, hotline, or exchange policy, inquire about washing temperature.
  \item \textbf{2: Technical Support} – If license issues arise for WX3510/WX5540, answer related questions, confirm email and phone, send manuals, ask for model and associated phone.
  \item \textbf{3: Financial Services – Pension} – If retirement certification reminder, say not remembered, will handle via WeChat, thank them, ask about past communication and WeChat face verification method.
  \item \textbf{4: Life Services} – If laundry fee update, respond with satisfaction, hope for discount, mention night service needs, ask about pricing explanation, training quality, and monthly discounts.
  \item \textbf{5: Financial Services – Trade Notice} – If transaction confirmation, say user unavailable. If medium-risk, ask for analyst opinion, arrange follow-up next day, inquire about transaction fees.
\end{itemize}
\end{tcolorbox}

\begin{tcolorbox}[
  colback=gray!4!white,
  colframe=gray!85!black,
  title=\textbf{Caller History and New Input},
  fonttitle=\bfseries,
  boxrule=1pt,
  top=3mm,
  bottom=3mm
]
\textbf{Conversation History}: \textit{(Empty)}

\textbf{Caller New Input}: Hello, this is the Forex Trading Customer Service Center. We are confirming that the currency pair you traded is USD/SGD, and the rate information is ready.
\end{tcolorbox}

\begin{tcolorbox}[
  colback=teal!2!white, 
  colframe=teal!80!black, 
  coltitle=white,
  title=\textbf{Model Response},
  fonttitle=\bfseries,
]
\textbf{Target Domain}: \textit{5: Financial Services – Trade Notice}

\textbf{Dialogue State}: \textit{Ongoing}

\textbf{Response}: Hello, this call is currently being handled by the smart assistant. The user is temporarily unavailable. If this is a medium-risk transaction, could we get advice from a market analyst? Please contact us again tomorrow for confirmation.
\end{tcolorbox}

\caption{A multi-turn example from the Honor-Dialogue dataset. The dataset features realistic, task-oriented, multi-domain conversations, in which each model response includes structured states. This example represents the situation of the first conversation.}
\label{fig:dialogue-example}
\end{figure*}

\subsubsection{Training Configuration}

In majority voting sampling, 5 samples are used, and data with more than 2 votes are filtered, with a sampling temperature of 0.7. The loss weights are set as $w_{\text{risk}}=0.3$, $w_{\text{CVaR}}=0.5$, $w_{\text{gain}}=0.3$, $w_{\text{shift}}=0.2$, $w_{\text{shape}}=0.1$, $w_{\text{curv}}=0.2$, and $w_{\text{consist}}=0.1$.
For the value model, the risk threshold is set to 0.1, the number of interval quantile points is 200, and the number of distributional value heads is 3. The reinforcement learning training runs for 1 iteration. During training, the model generates dialogue tasks of length 4096 for reasoning tasks and 1024 for dialogue tasks. The experiments are conducted on 8 × NVIDIA A100 80GB GPUs.

\subsubsection{Dialogue Task Evaluation}
Our dialogue evaluation leverages GPT-4o, but the model is not utilized as a free-form judge. Instead, it implements a rigorous rubric-based evaluation protocol designed to mitigate subjectivity and enhance reproducibility. This rubric explicitly defines evaluation metrics, adopts a 1–5 scoring scale with clear criteria, includes two scoring examples and a standardized output format, and the prompt incorporates a well-structured process along with comprehensive dialogue content and contextual information. GPT-4o functions solely as an automated evaluator applying this fixed rubric, rather than an unconstrained scorer.

To further ensure reliability, we conducted cross-validation through two key steps: \textbf{1)} multiple sampling runs to verify the stability of rubric execution, and \textbf{2)} spot-checked human evaluations that demonstrated high agreement with rubric-based scores. Specifically, for the comparative experiment between human annotators and the model, we validated the protocol using 1,110 data samples covering 10 scenarios. The statistical details of the validation dataset are presented in Table \ref{tab:validation-data-statistics}.

Four independent annotators scored these samples strictly in accordance with the rubric. They are professional data annotators in the company. Table \ref{tab:human-model-agreement} presents the experimental results for goal completion rate scoring (full score: 5) in dialogue tasks.
As shown in Table \ref{tab:human-model-agreement}, the mean score assigned by human annotators is extremely close to that of the model, with a difference of only 0.021 points. These findings confirm strong consistency between human and model scores, validating the effectiveness of our rubric. We acknowledge that fully establishing external validity necessitates additional independent annotators. We are in the process of releasing our rubric, evaluation prompts, and real-dialogue dataset to facilitate replication.
To illustrate the evaluation logic for dialogue performance, the core prompt section regarding the assessment of conversation logicality is provided as follows in Figure\ref{fig:scoring-guidelines}, which is a part of goal completion rate.
\begin{figure*}[!t]  
\centering

\begin{tcolorbox}[
  colback=gray!4!white,
  colframe=gray!85!black,
  title=\textbf{Core Prompt for Dialogue Logicality},
  fonttitle=\bfseries,
  boxrule=1pt,
  top=3mm,
  bottom=3mm
]
When scoring the intelligent call assistant's responses, please follow these steps:

1.Read and understand the entire call content.

2.Score from the dimension of dialogue logic rationality. The definition of dialogue logic rationality is: whether all responses provided by the assistant during the entire dialogue are clear, coherent, and effectively convey relevant information. A reasonable response should ensure the consistency, coherence, and relevance of information. Use a 1-5 scoring scale with specific criteria as follows:
\begin{itemize}
    \item \textbf{1 point:} The response exhibits severe logical inconsistencies and violates fundamental reasoning principles. The intent of the assistant is unclear, and the response fails to convey meaningful or usable information.

    \item \textbf{2 points:} The response contains multiple inconsistencies or conflicting statements. Although partially related to the query, it lacks sufficient contextual grounding and coherent reasoning, resulting in poor interpretability and weak informational value.

    \item \textbf{3 points:} The response is generally coherent but contains minor logical gaps or discontinuities that affect fluency. While relevant to the query, the explanation lacks depth or clarity in parts, limiting the overall effectiveness of communication.

    \item \textbf{4 points:} The response is logically consistent and well-structured, with clear and relevant content. Information is conveyed effectively and aligns well with the user’s intent, enabling smooth and coherent interaction.

    \item \textbf{5 points:} The response demonstrates strong logical consistency, clear reasoning, and precise information delivery. It not only addresses the query accurately but also facilitates deeper understanding, effectively guiding the conversation and enhancing overall interaction quality.
\end{itemize}

3.Based on the above scoring criteria and combined with the entire dialogue content, give a reasonable score, along with the scoring thinking process, deduction points, and modification suggestions to improve the score without changing the original meaning of the text.
\end{tcolorbox}

\caption{A core prompt for dialogue logicality assessment from the constructed rubric evaluation method. The rubric features clear scoring steps, explicit definition of logical rationality, and detailed 1-5 point end-point criteria, in which each evaluation requires supplementary scoring reasoning, deduction explanations, and optimization suggestions. }
\label{fig:scoring-guidelines}
\end{figure*}

\subsection{Additional experimental details in the Dialogue Task}

Table~\ref{tab:dialog-full} presents the comprehensive results across all five real-world dialogue domains. DVPO achieves the highest overall performance with an average score of 86.79\%, outperforming all RL baselines. Specifically, DVPO surpasses PPO and Reinforce++ by 1.59\% and 2.94\% on average, and exceeds GRPO-based methods by more than 20 percentage points in most domains. This advantage is consistent across all metrics: task completion, ask completion, and goal completion.

A detailed analysis of domain-level results reveals the stability of DVPO under in domain and out of
domain tasks. In Life Services and Transportation \& Travel, DVPO achieves domain averages exceeding 88\%, outperforming PPO by approximately 6 percentage points and Reinforce++ by 3--5 percentage points. In the more challenging Healthcare \& Wellness and Social \& Entertainment domains, DVPO maintains an accuracy of 87--88\%, whereas PPO exhibits fluctuations of up to 10 percentage points and GRPO deteriorates to the 30--50\% range. Even in the most demanding Financial Services domain, DVPO remains above 85\%, limiting the performance drop to within 3 percentage points, while GRPO variants degrade by more than 20 percentage points.

The cross-domain results further underscores DVPO’s robustness and generalization.These results demonstrate that DVPO provides a reliable and noise-resilient reinforcement learning strategy for real-world dialogue systems with imperfect supervision.

\subsection{Additional Ablation Experiment}

\paragraph{DVPO exhibits strong scalability across model scales}
As shown in Table~\ref{tab:small-model}, DVPO consistently outperforms PPO, GRPO, and Reinforce++ even when applied to the smaller Qwen3-1.7B model.  
Despite the reduced capacity, it achieves an average accuracy of $58.22\%$ on mathematical tasks and $2.82\%$ on scientific tasks, maintaining stable optimization under noisy rewards.  
These results confirm that DVPO remains effective across different model scales, demonstrating robust generalization and adaptability under noisy training conditions.

\paragraph{Effectiveness of Distributional Constraints on Robustness and Generalization}

Table~\ref{tab:dvpo-ablation} presents an ablation study of different distributional loss components in DVPO.
Starting from the core quantile regression baseline, we observe that introducing additional distributional constraints consistently improves performance across most settings.
Among all components, tail-related objectives contribute the most significant gains.
While distribution consistency and tail calibration improve robustness to some extent, their effects are more pronounced on out-of-domain tasks, especially on challenging math benchmarks such as AIME24 and AMC23.
Notably, incorporating tail shape and curvature modeling leads to the most consistent improvements across both in-domain and out-of-domain evaluations, achieving the highest overall accuracy of 39.63\%.
These results indicate that explicitly modeling the shape of the value distribution plays a critical role in improving generalization.
Compared to variance-reduction or calibration-based objectives alone, shape-aware constraints provide stronger structural regularization, enabling DVPO to better capture distributional properties under noisy or shifted reward settings.

\subsection{Additional Visualization}

\begin{figure*}[h]
    \centering
    \includegraphics[width=1\textwidth]{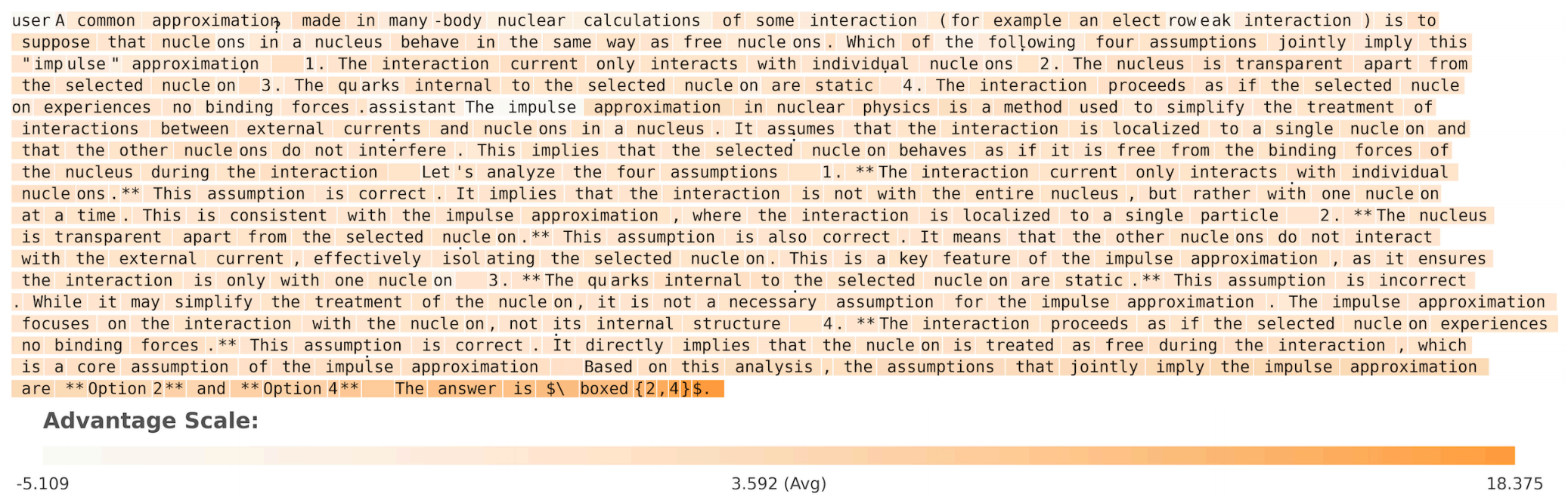}
    \caption{Advantage estimation visualization of the PPO method for the target question output. The method correctly identifies the final output as erroneous, while the token-level advantage values show minimal variability.}
    \label{fig:ppo-w}
\end{figure*}

\begin{figure*}[h]
    \centering
    \includegraphics[width=1\textwidth]{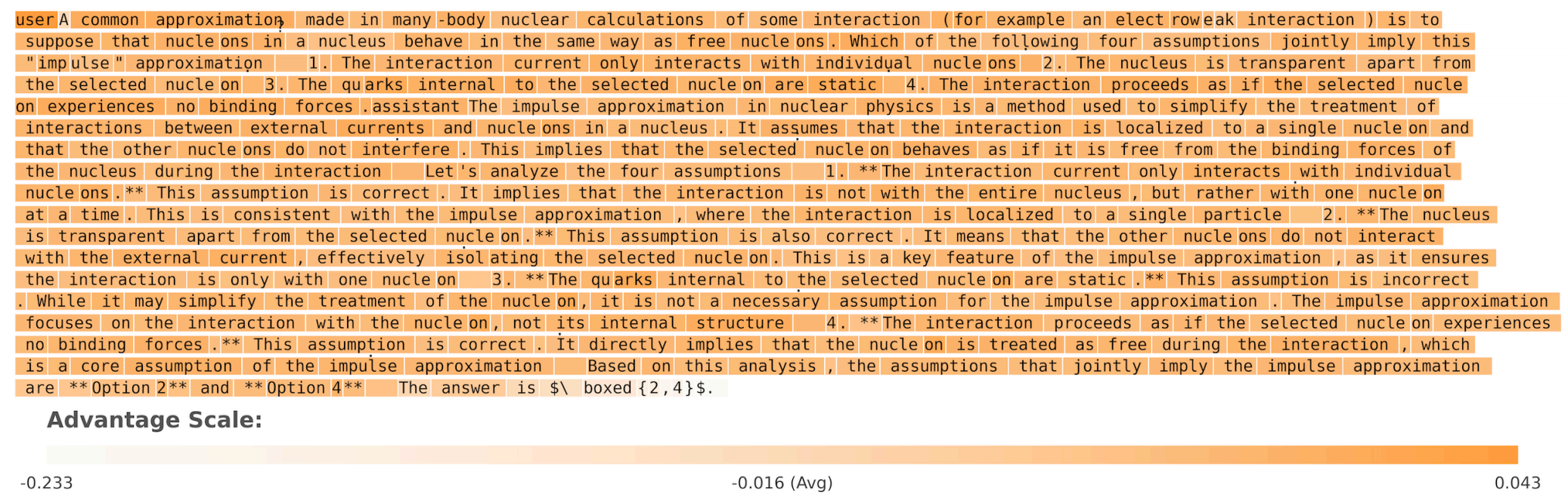}
    \caption{Advantage estimation visualization of the Robust Bellman PPO method for the target question output. The method fails to recognize the invalidity of the novel output, with negligible differences in token-level advantage estimations and overestimation of the final output's advantage.}
    \label{fig:bei-w}
\end{figure*}

\begin{figure*}[h]
    \centering
    \includegraphics[width=1\textwidth]{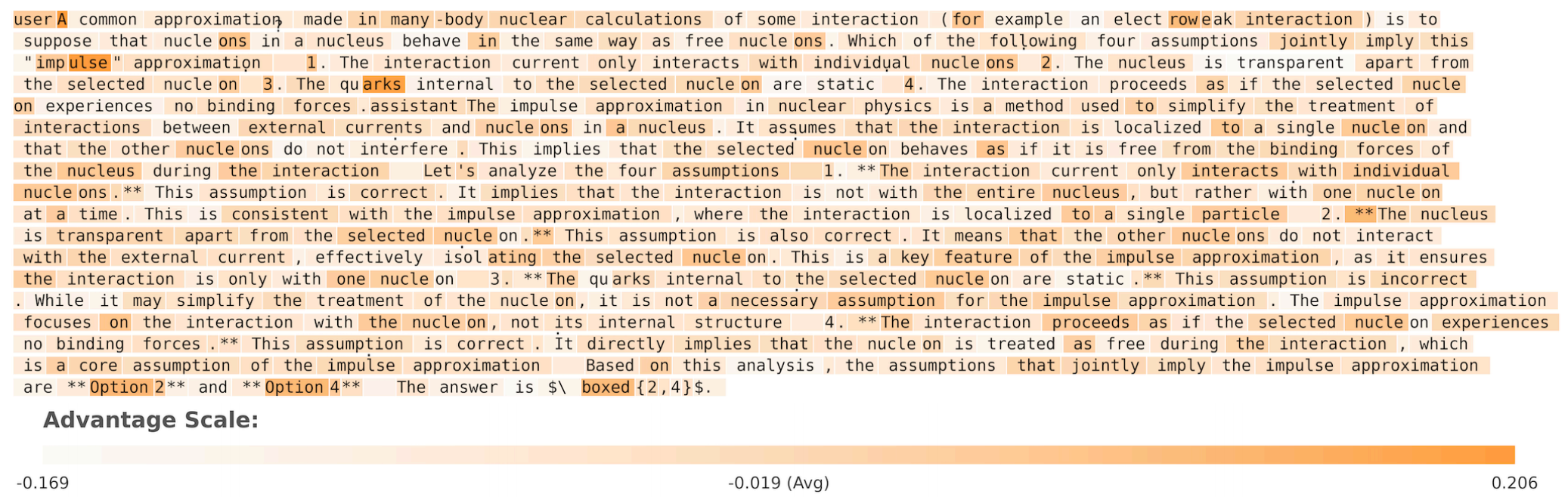}
    \caption{Advantage estimation visualization of our proposed DVPO method for the target question output. The method accurately judges the correctness of the final output and effectively captures key semantic terms (e.g., "nucleus" and "quarks") via distinct token-level advantage weights.}
    \label{fig:our-w}
\end{figure*}

Figures~\ref{fig:ppo-w}, \ref{fig:bei-w}, and \ref{fig:our-w} illustrate a comparative analysis of advantage estimations among PPO, Robust Bellman PPO, and our proposed method (DVPO) for the same question outputs, irrespective of their semantic quality.
Notably, significant discrepancies in advantage estimation patterns are observed across the three methods. For PPO, despite correctly identifying the final output as erroneous, the advantage values assigned to individual tokens exhibit minimal variability. Similarly, Robust Bellman PPO fails to recognize the invalidity of the novel output and also demonstrates negligible differences in token-level advantage estimations. Additionally, it overestimates the advantage of the final output component. This observation partially reflects the limited generalization capability of these two methods when handling novel response generations.

In contrast, our DVPO method accurately judges the correctness of the final output. More importantly, its advantage estimation effectively captures key semantic terms (e.g., "nucleus" and "quarks") by assigning distinct advantage weights to these critical tokens, which aligns with the core semantic characteristics of the target task.

\end{document}